\documentclass{new_tlp}
\usepackage{mathptmx}
\usepackage{color}

\usepackage{listings}
\usepackage{xcolor}
\definecolor{codegreen}{rgb}{0,0.6,0}
\definecolor{codegray}{rgb}{0.5,0.5,0.5}
\definecolor{codepurple}{rgb}{0.58,0,0.82}
\definecolor{backcolour}{rgb}{0.95,0.95,0.92}

\lstdefinestyle{mystyle}{
    backgroundcolor=\color{backcolour},   
    commentstyle=\color{codegreen},
    keywordstyle=\color{magenta},
    numberstyle=\tiny\color{codegray},
    stringstyle=\color{codepurple},
    basicstyle=\ttfamily\footnotesize,
    breakatwhitespace=false,         
    breaklines=true,                 
    captionpos=b,                    
    keepspaces=true,                 
    numbers=left,                    
    numbersep=5pt,                  
    showspaces=false,                
    showstringspaces=false,
    showtabs=false,                  
    tabsize=2
}

\newcommand\bcmdtab{\noindent\bgroup\tabcolsep=0pt%
  \begin{tabular}{@{}p{10pc}@{}p{20pc}@{}}}
\newcommand\ecmdtab{\end{tabular}\egroup}

\usepackage{amsmath}
\usepackage{graphicx}
\usepackage{multirow}
\usepackage{listings}

\usepackage[T1]{fontenc}
\usepackage{xspace}
\usepackage{comment}
\usepackage{hyperref}
\usepackage[mathscr]{euscript}

\def\aopl{$\mathscr{AOPL}$\xspace}
\def\aoplprime{$\mathscr{AOPL'}$\xspace}
\def\aoplp{$\mathscr{AOPL}$-$\mathscr{P}$\xspace}

\def\ald{$\mathscr{AL}_d$\xspace}
\def\p{$\mathscr{P}$\xspace}
\def\t{$\mathscr{T}$\xspace}
\def\lp{$lp(\mathscr{P}, \sigma)$\xspace}
\def\Ps{$\mathscr{P}(\sigma)$\xspace}

\def\reilp{$rei\_lp(\mathscr{P})$\xspace}

\def\ev{$\langle \sigma, ca \rangle$\xspace}
\def\evi{$\langle \sigma_i, ca_i \rangle$\xspace}
\def\e{$\mathscr{E(P)}$\xspace}
\def\r{$\mathscr{R}$\xspace}
\def\et{$\mathscr{E}(t)$\xspace}

\newtheorem{example}{Example}
\newtheorem{definition}{Definition}
\newtheorem{proposition}{Proposition}

  \title[Autonomous Agents \& Policy Compliance: A Framework for Reasoning About Penalties]
        {Autonomous Agents and Policy Compliance: A Framework for Reasoning About Penalties}

  \author[V. Tummala and D. Inclezan]
         {VINEEL TUMMALA and DANIELA INCLEZAN\\
         Miami University, Oxford, OH, USA\\
         \email{tummalvs@miamioh.edu, inclezd@miamioh.edu}}

\jdate{May 2023}
\pubyear{2023}
\pagerange{\pageref{firstpage}--\pageref{lastpage}}

\begin{document}

\label{firstpage}

\maketitle       

\begin{abstract}
This paper presents a logic programming-based framework for policy-aware autonomous agents that can reason about potential penalties for non-compliance and act accordingly. 
While prior work has primarily focused on ensuring compliance, our approach considers scenarios where deviating from policies may be necessary to achieve high-stakes goals. Additionally, modeling non-compliant behavior can assist policymakers by simulating realistic human decision-making.
Our framework extends Gelfond and Lobo's  Authorization and Obligation Policy Language (\aopl) to incorporate penalties and integrates Answer Set Programming (ASP) for reasoning. Compared to previous approaches, our method ensures well-formed policies, accounts for policy priorities, and enhances explainability by explicitly identifying rule violations and their consequences. Building on the work of Harders and Inclezan, we introduce penalty-based reasoning to distinguish between non-compliant plans, prioritizing those with minimal repercussions. To support this, we develop an automated translation from the extended \aopl into ASP and refine ASP-based planning algorithms to account for incurred penalties.
Experiments in two domains demonstrate that our framework generates higher-quality plans that avoid harmful actions while, in some cases, also improving computational efficiency. These findings underscore its potential for enhancing autonomous decision-making and informing policy refinement.
Under consideration in Theory and Practice of Logic Programming (TPLP).
\end{abstract}


\section{Introduction}

In this paper we explore autonomous agents operating in dynamic environments governed by \textit{policies} or \textit{norms}, including cultural conventions and regulations. Our focus is on agents whose knowledge bases and reasoning algorithms are encoded in logic programming, specifically Answer Set Programming (ASP) \cite{gl91,mt99}. We introduce a framework that enables these policy-aware agents to assess potential penalties for non-compliance and generate suitable plans for their goals. 

Research on norm-aware autonomous agents predominantly focuses on compliance (e.g., \cite{Oren11,Alechina12}). However, studying non-compliant agents is equally important for two key reasons.  
First, autonomous agents may be tasked with high-stakes objectives (e.g., assisting in rescue operations) that can only be achieved through selective non-compliance. In such cases, identifying optimal non-compliant plans that accomplish the goal while minimizing repercussions is crucial.  
Second, our framework can support policymakers by enabling policy-aware agents to model human behavior. Humans do not always adhere to norms and often seek to minimize penalties for non-compliance. By simulating different compliance attitudes, policymakers can identify potential weaknesses in policies and refine them accordingly.

To enable autonomous agents to reason about policies, evaluate penalties for non-compliance, and generate optimal plans based on their circumstances and given norms, we must first encode these policies. In our proposed framework, policies are specified in the \textit{Authorization and Obligation Policy Language} (\aopl) by Gelfond and Lobo \citeyear{gl08}, whose semantics are defined via a translation into ASP. We expand \aopl to enable the representation of, and reasoning about, penalties that may be incurred for non-compliance with a policy.
We develop an automated translation of the extended \aopl, referred to as \aoplp, into ASP and refine ASP-based planning algorithms to incorporate penalty considerations.

Using a high-level language such as \aopl for representing policies and penalties offers several advantages over alternative approaches, which may include encoding policies as soft constraints in ASP \cite{Calimeri2020}. One key benefit is that \aopl ensures the policy is well-structured, with built-in mechanisms to detect issues like inconsistencies, ambiguities, and underspecification \cite{di23}. Additionally, \aopl’s syntax and semantics support the representation of priorities between policy statements, allowing for the determination of which statements apply at different points in time. Moreover, our \aopl-based approach enhances explainability by identifying which policy rules an agent violated and how these violations led to accumulated penalties---something that would be difficult to achieve using soft constraints.

In previous work on policy-aware autonomous agents, Harders and Inclezan \citeyear{hi23} introduced \textit{behavior modes} to capture different attitudes toward policy compliance, allowing non-compliant actions within the Risky mode. While this framework established a foundation for reasoning about non-compliance, it ranked plans in the Risky mode solely by their length, without accounting for the number or proportion of non-compliant actions or the severity of the violated rules. However, other behavior modes---excluding non-compliant ones---did consider the proportion of explicitly known compliant actions, for example.

In this work, we extend that approach by introducing a more nuanced evaluation of non-compliance, enabling agents to weigh trade-offs between policy violations and their objectives more effectively. Our framework distinguishes between non-compliant plans using penalties, allowing agents to achieve their goals while minimizing repercussions. Experimental results show that our approach generates higher-quality plans. For example, in a Traffic Norms domain, it selects an optimal driving speed and avoids actions potentially harmful to humans. Such factors are overlooked by the previous framework, which accounts only for plan length while ignoring execution time (and thus driving speed) and gives no special consideration to preventing human harm.
 Additionally, for certain domains, our method demonstrates improved efficiency. A preliminary version of this work appears in \cite{TummalaI24}.

The rest of the paper starts with background information in Section \ref{sec:background}, followed by a motivating example in Section \ref{sec:example}. We present our framework in Section \ref{sec:framework}, which includes the extension of \aopl with penalties (\aoplp), an automated translator from \aoplp to ASP, penalty-aware planning, and revised behavior modes.
 We present experimental results in Section \ref{sec:experimental_resuts}. We discuss related work in Section \ref{sec:related_work} and end with conclusions and future work in Section \ref{sec:conclusions}.

\section{Background}
\label{sec:background}
We now provide background on the norm-specification language \aopl and our previous work on policy-aware planning agents. We assume readers are familiar with ASP or can consult external resources (e.g., \cite{gl91,gk14,Calimeri2020}) as needed.

\subsection{Policy-Specification Language \aopl}
\label{sec:aopl}
Gelfond and Lobo \citeyear{gl08} 
designed the Authorization and Obligation Policy Language \aopl for specifying policies for an intelligent agent acting in a dynamic environment. A policy is a collection of authorization and obligation statements.  
An {\em authorization} indicates whether an agent's action is permitted or not, and under which conditions. 
An {\em obligation} specifies whether an agent is required or not required to perform a particular action or to abstain from it under given conditions. 
An \aopl policy assumes that the agent's environment is described using an action language, a high-level language designed to concisely and accurately represent action preconditions as well as the direct and indirect effects of actions. Over time, various action languages have been developed (e.g., $\mathscr{A}$ \cite{Gelfond1993}, $\mathscr{B}$ \cite{Gelfond1998}, $\mathscr{C+}$ \cite{Giunchiglia2004}, $\mathscr{H}$ \cite{Chintabathina2005}, \ald \cite{gi13}, etc.), including modular action languages such as MAD \cite{Lifschitz2006} and $\mathscr{ALM}$ \cite{Inclezan2012}. They all incorporate means for dealing with the ramification and qualification problems, as well as the law of inertia (i.e.,  domain properties not affected by actions remain unchanged).
 
 A system description written in an action language defines the domain's transition diagram 
whose states are complete and consistent sets of static (i.e., immutable) and fluent literals (i.e., properties of the domain that may be changed by actions), and whose arcs are labeled by actions.
The signature of the dynamic system description (which includes predicates denoting {\em sorts}, {\em statics}, {\em fluents}, and {\em actions}) is included in the signature of an \aopl policy for that dynamic domain. 
Additionally, the signature of an \aopl policy includes predicates $permitted$ for authorizations,
$obl$ for obligations, and \textit{prefer} for specifying preferences between authorizations or between obligations. A \textit{prefer} atom is created from the predicate \textit{prefer}; similarly, for $permitted$ and $obl$ atoms. A literal is an atom or its negation.

An \aopl \textit{policy} \p is a finite collection of statements of the form:
\begin{subequations} \label{eq2}
    \begin{align}
            & \ \ permitted\left(e\right) & \textbf{ if } \ cond \label{eq1_1}\\[-0.3em]
            & \neg permitted\left(e\right) & \textbf{ if } \ cond \label{eq1_2}\\[-0.3em]
		& \ \ obl\left(h\right) & \textbf{ if } \ cond \label{eq1_3}\\[-0.3em]
            & \neg obl\left(h\right) & \textbf{ if } \ cond\label{eq1_4}\\[-0.3em]
        d: \textbf{normally } & \ \ permitted(e) & \textbf{ if } \ cond \label{eq2_1}\\[-0.3em]
        d: \textbf{normally } & \neg permitted(e) & \textbf{ if } \ cond \label{eq2_2}\\[-0.3em]
        d: \textbf{normally } & \ \ obl(h) & \textbf{ if } \ cond \label{eq2_3}\\[-0.3em]
        d: \textbf{normally } & \neg obl(h) & \textbf{ if } 
        \ cond \label{eq2_4}\\[-0.3em]
	    	& \ \ \textit{prefer}(d_i, d_j) & \label{eq2_5}
    \end{align}
\end{subequations}
where $e$ is an elementary action; $h$ is a happening (i.e., an elementary action or its negation\footnote{If $obl(\neg e)$ is true, 
then the agent must not execute $e$.}); 
$cond$ is a 
set of literals of the signature, except for \textit{prefer} literals; $d$ appearing in (\ref{eq2_1})-(\ref{eq2_4}) denotes a defeasible rule label; and $d_i$, $d_j$ in (\ref{eq2_5}) refer to distinct \emph{defeasible} rule labels from \p.
Rules (\ref{eq1_1})-(\ref{eq1_4}) encode {\em strict} policy statements, while rules (\ref{eq2_1})-(\ref{eq2_4}) encode {\em defeasible} statements (i.e., statements that may have exceptions). Rule (\ref{eq2_5}) captures \textit{priorities} between defeasible statements only. It specifies that a defeasible rule labeled $d_i$ overrides a defeasible rule labeled $d_j$, rendering the latter inapplicable when the condition of the former (i.e., of $d_i$) is satisfied. Strict rules, whenever their condition $cond$ is satisfied, always override the defeasible rules they conflict with.
Unlike deontic logic, the \aopl language described by Gelfond and Lobo does not assume an equivalence between rules for $\neg permitted(e)$ and  $obl(\neg e)$. We believe this choice was made to allow for different interpretations and to accommodate other types of logics. Such an equivalence can be implemented by adding the following rules to an \aopl policy \p:
$$
\begin{array}{lll}
\neg permitted(e) & \textbf{if} & obl(\neg e)\\
obl(\neg e) & \textbf{if} & \neg permitted(e)
\end{array}
$$
However, the lack of a clear relationship between permissions and obligations can give rise to \textit{modality conflicts}, a term introduced by Craven et al. \citeyear{clmrlb09}, as noted by Inclezan \citeyear{di23}. For example, an \aopl policy may derive both $\neg permitted(e)$ and $obl(e)$ for the same elementary action $e$ in a given state, which is an undesirable outcome, as it appears contradictory.

The semantics of an \aopl policy determine a mapping \Ps from states of a transition diagram \t into a collection of $permitted$ and $obl$ literals, obtained from the policy statements that are applicable in state $\sigma$ (i.e., have a satisfied $cond$ and are not overridden). To formally describe the semantics of \aopl, a translation of a policy \p and a state $\sigma$ of the transition diagram into ASP is defined as \lp. 
Properties of an \aopl policy \p are defined in terms of the answer sets of the logic program \lp expanded with appropriate rules. 
Gelfond and Lobo define a policy as \textit{consistent} if, for every state $\sigma$ of \t, the logic program \lp is consistent (i.e., has an answer set). A policy is \textit{categorical} if \lp has \textit{exactly one answer set} for every state $\sigma$ of \t. 

In our work, we adopt established definitions for classifying events as strongly-compliant (i.e., explicitly permitted), underspecified (i.e., neither explicitly permitted nor explicitly prohibited), or non-compliant (i.e., explicitly prohibited) with respect to authorizations, and as compliant or non-compliant with respect to obligations.
Formal definitions for these concepts, adapted from Gelfond and Lobo \citeyear{gl08} and Inclezan \citeyear{di23}, are provided below. Note that $ca$ denotes a compound action, while $e$ refers to an elementary action. An event \ev is a pair consisting of a state $\sigma$ and a (possibly compound) action $ca$ occurring in that state.
If $l$ is a literal, then \lp $\models l$, read as ``the logic program \lp entails $l$,'' denotes that $l$ belongs to every answer set of \lp. Similarly, \lp $\not\models l$ read as ``the logic program \lp does not entail $l$,'' denotes that $l$ does not appear in any of the answer sets of \lp.

\begin{definition}[Compliance for Authorizations] 

\label{def:auth_compliance}
\begin{itemize}
\item An event \ev is {\em strongly-compliant} with respect to the authorizations in policy \p if, for every $e \in ca$, we have that \lp $\models permitted(e)$.
\item An event \ev is {\em underspecified} with respect to the authorizations in policy \p if, for every $e \in ca$, we have that 
\lp $\not\models permitted(e)$ and
\lp $\not\models \neg permitted(e)$.
\item An event \ev is {\em non-compliant} with respect to the authorizations in policy \p if, for every $e \in ca$, we have that
\lp $\models \neg permitted(e)$.
\end{itemize}
\end{definition}


\begin{definition}[Compliance for Obligations]
\label{def:obl_compliance}

An event \ev is {\em compliant} with respect to the obligations in policy \p if 

\noindent
$\bullet\ $ For every $e$ such that \lp $\models obl(e)$ we have that $e \in ca$, and

\noindent
$\bullet\ $ For every $e$ such that \lp $\models obl(\neg e)$ we have that $e \notin ca$.
\end{definition}


\begin{definition}[Compliance for Authorization and Obligations]
\label{def:auth_obl_compliance}
An event \ev is {\em strongly compliant} with arbitrary policy \p (that may contain both authorizations and obligations) if it is strongly compliant with the authorization component and compliant with the obligation component of \p.
\end{definition}

Inclezan \citeyear{di23} showed that in categorical (i.e., unambiguous) policies, events \ev such that $ca$ consists of a single elementary action 
can be categorized with respect to authorization as either strongly-compliant (i.e., explicitly permitted), non-compliant (i.e., explicitly prohibited), or underspecified (i.e., neither explicitly permitted nor explicitly prohibited).
In the case of non-categorical consistent policies, there will be events that lie outside of this categorization (i.e., do not fit in any of these three categories). 
A typical example of a non-categorical consistent policy is one that consists of two defeasible rules, both applying in a state $\sigma$, one deriving $permitted(e)$ and the other deriving $\neg permitted(e)$ with no preference stated between them. In such a case, \lp has two answer sets where one answer set contains $permitted(e)$, while the other contains $\neg permitted(e)$; hence none of the categories in Definition~\ref{def:auth_compliance} apply to event $\langle \sigma, e \rangle$.
Similarly, in the case of compound actions (i.e., when the agent executes more than one action at a time). A compound action may include elementary actions belonging to different categories (e.g., one compliant and another underspecified), and thus the compound action itself does not fit into any single category.

\textit{In what follows, we assume categorical policies.} Handling non-categorical policies is non-trivial and left for future work. However, a well-defined policy is generally expected to be categorical, while a non-categorical (i.e., ambiguous) policy typically reflects a flaw in its specification or design.

\subsection{Agent Behavior Modes with Respect to Policy Compliance}
\label{sec:HI_framework}
In previous work, Harders and Inclezan \citeyear{hi23} introduced an ASP framework for plan selection in policy-aware autonomous agents, where policies were specified in \aopl. They proposed that agents could adopt different attitudes toward norm compliance, influencing the selection of the ``best'' plan. These attitudes, termed \textit{behavior modes}, were defined using various metrics to capture different compliance strategies. These metrics included plan length as well as the number and percentage of different types of \textit{elementary} actions: strongly compliant actions (explicitly permitted), underspecified actions (neither explicitly permitted nor prohibited), and non-compliant actions (violating authorizations or obligations).
 
The following predefined agent behavior modes were introduced by Harders and Inclezan:
\begin{itemize}
    \item Safe Behavior Mode -- prioritizes  actions that are explicitly known to be compliant (i.e., maximizes the percentage of strongly-compliant elementary actions first and then plan length) and does not execute non-compliant actions;
\item Normal Behavior Mode -- prioritizes plan length and then actions explicitly known to be compliant (i.e., minimizes plan length first and then maximizes the percentage of strongly-compliant elementary actions), while not executing non-compliant actions; and 
\item Risky Behavior Mode -- disregards policies, but does not go out of its way to be non-compliant either. This may result in the inclusion of non-compliant actions if they contribute to minimizing plan length.
\end{itemize}

Harders and Inclezan \citeyear{hi23} encoded these behavior modes in the ASP-variant that constitutes the input language for the \textsc{Clingo} solver \cite{clingo2019} using constraints, employing the $\#maximize$ and $\#minimize$ constructs to express priorities. For instance, in the Safe behavior mode, the planning module included the following rules:

$
\begin{array}{l}
\#maximize\ \{N@2\ :\  p\_sa(N)\}\\
\#minimize\ \{N@1\ :\ l(N)\}\\
\leftarrow n\_na(N), \mbox{ not } N = 0\\
\leftarrow n\_no(N), \mbox{ not } N = 0
\end{array}
$

\noindent
where $p\_sa$ is the percentage of strongly-compliant elementary actions (with respect to authorizations); $l$ is the length of the plan; $n\_na$ and $n\_no$ denote the number of non-compliant elementary actions with respect to authorizations and obligations, respectively. 

\section{Motivating Example: Traffic Norms Domain}
\label{sec:example}

\begin{figure} [htb]
    \centering
    \includegraphics[width = 0.7\textwidth]{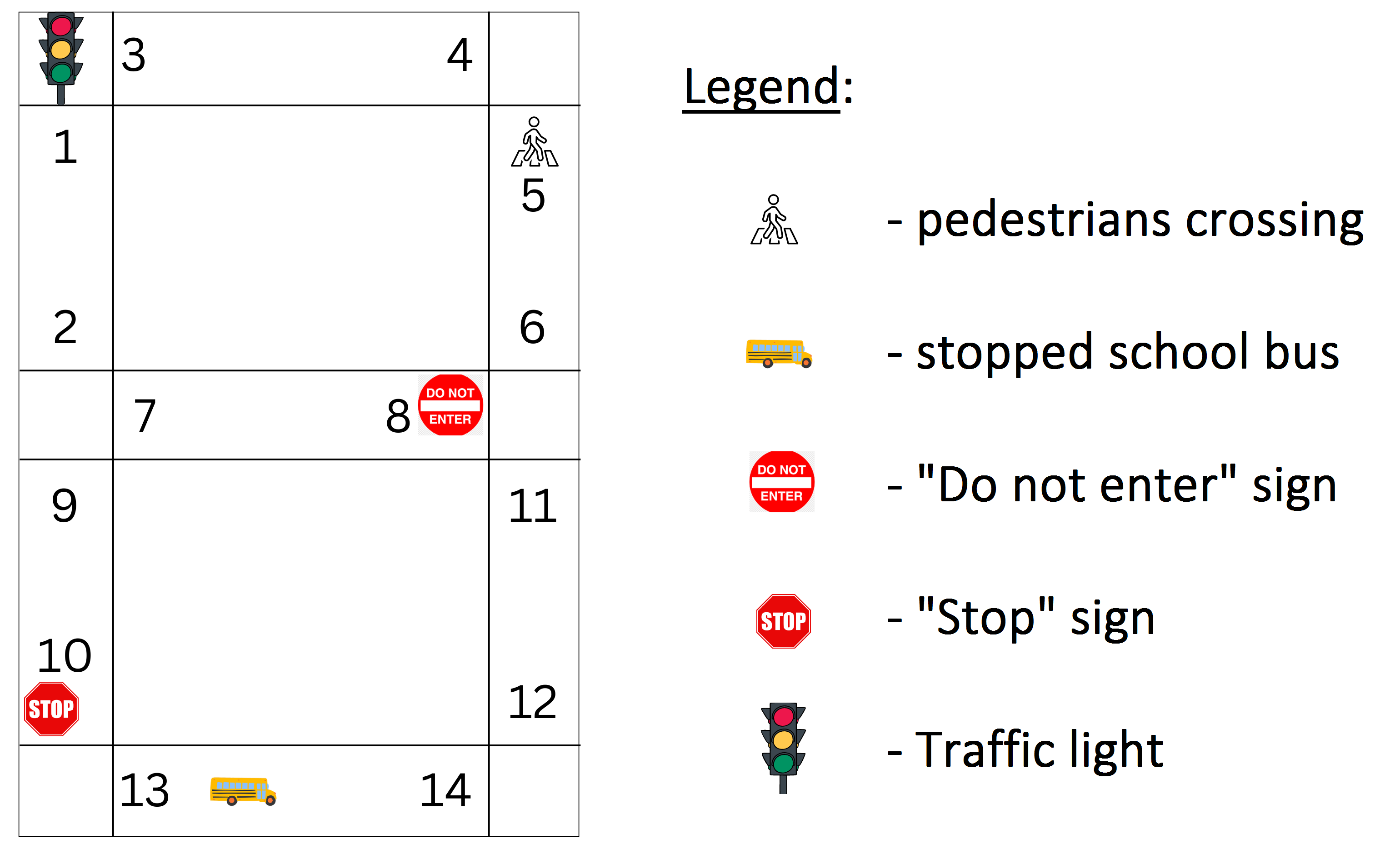}
    \caption{Layout of the Traffic Norms Domain }
    \label{fig:self_driving agent}
\end{figure}

To illustrate the need for penalties in enabling more nuanced planning, especially allowing certain levels of non-compliance in high-stakes situations, consider a dynamic domain where a self-driving agent navigates a simplified city environment. There are certain norms that the self-driving agent must be aware of when driving, which may represent traffic regulations or cultural conventions. We limit ourselves to one agent, a few traffic signs, and a grid street layout. A schematic view of this \textit{Traffic Norms Domain} is in Figure~\ref{fig:self_driving agent}.

To model this dynamic domain, we consider 
fourteen locations labeled from $1$ to $14$; a set of driving speeds; 
traffic light colors red, yellow, and green; and two traffic signs ``Stop'' and ``Do not enter.'' 
One example of a fluent in this domain is $pedestrians\_are\_crossing(L)$ saying that people are crossing the street at location $L$. 
The agent can execute actions: $drive(L_1, L_2, S)$ to drive between two (connected) locations $L_1$ and $L_2$ at a speed $S > 0$; and $stop(L)$ to stop at $L$.

We consider the set of policies in Figure \ref{fig:policies}.  
Note that some of these rules represent cultural norms (e.g., rules 1 and 2), while others reflect traffic regulations (e.g., rule 5). For illustration purposes, we consider a system of penalties on a scale from 1 to 3 and discuss potential issues with this choice in Section \ref{sec:discussion}. In practice, such penalties would be determined by experts in ethics and traffic regulation; here, we adopt a 3-point scale solely to illustrate the framework. Policy rules 7 and 8 do not have penalties associated with them, as they describe permissions. 

\small
\begin{figure}[hbt]
\noindent\fbox{%
    \parbox{0.98\textwidth}{%
\begin{enumerate}

\item The agent is constrained by a defeasible rule that does \textit{not permit} exceeding the speed limit by more than 5 mph if the speed limit is under 55 mph. 

\textit{Penalty:} 1-point penalty if exceeding the speed limit by less than 10 mph; 2-point if exceeding the speed limit by 10 to 19 mph; and a 3-point penalty is imposed if exceeding the speed limit by 20 mph or more.

\item The agent is constrained by a defeasible rule that does \textit{not permit} exceeding the speed limit by more than 10 mph when the speed limit is 55 mph or higher. 

\textit{Penalty:}  a 2-point penalty if exceeding the speed limit by 10 to 19 mph, and a 3-point penalty if exceeding the speed limit by 20 mph or more.

\item The agent is strictly \textit{obligated not} to enter the roads marked as ``Do not enter.'' 

\textit{Penalty:} 3-point penalty.

\item The agent is strictly \textit{not permitted} to roll over a stop sign. 

\textit{Penalty:} 2-point penalty.

\item The agent is strictly \textit{obligated not} to move when a school bus is stopped. 

\textit{Penalty:} 3-point penalty. (\textit{Note: we will refine this penalty in Section \ref{sec:discussion}.})

\item The agent is strictly \textit{obligated} to stop when pedestrians are crossing. 

\textit{Penalty:}  3-point penalty. (\textit{Note: we will refine this penalty in Section \ref{sec:discussion}.}) 

\item The agent is \textit{permitted} by a defeasible rule to proceed through an intersection when the traffic light turns green.

\item The agent is  \textit{permitted} by a defeasible rule to drive when the traffic light is yellow.

\item The agent is strictly \textit{obligated not} to cross an intersection on a red traffic light.

\textit{Penalty:} 3-point penalty.

\item Rule 1 overrides rules 7 and 8.

\item Rule 2 overrides rules 7 and 8.
\end{enumerate}
}%
}
\caption{Policies and Penalties for the Traffic Norms Domain}
\label{fig:policies}
\end{figure}
\normalsize

Harders and Inclezan \citeyear{hi23} introduced the Risky agent behavior mode for  
emergency situations.  
Agents in this mode only look for the shortest plan while ignoring policies. As a result, a Risky agent in the Traffic Norms domain who starts in location 6 and needs to get to location 1, where the speed limit between locations 2 and 1 is 45 mph, may come up with several plans (we don't specify speeds unless relevant to our discussion), including: 
\begin{enumerate}
    \item Drive from 6 to 5, from 5 to 4 without stopping for the pedestrians crossing the street, then from 4 to 3, and finally from 3 to 1 (possibly on a red light).
    \item Drive from  6 to 8 and thus enter a ``Do not enter'' street, drive from 8 to 7, from 7 to 2, and finally from 2 to 1 at 65 mph.
    \item Same as plan 2, but drive from 2 to 1 at 45 mph.
\end{enumerate}
According to the definition of Risky agents, all these plans are treated as equivalent, as they consider only plan length without factoring in the severity of infractions, such as the number or gravity of non-compliant actions.
Instead, we aim to distinguish between plans containing non-compliant actions and prioritize those that incur the least penalty. In scenarios where a fully compliant plan is unavailable, an agent may need to choose a non-compliant plan to achieve a high-stakes goal. To enable this, we introduce penalties as a means of guiding plan selection.

\section{Penalization Framework for Policy-Aware 
Agents}
\label{sec:framework}

In our framework, we assume that penalties for a domain are set by domain experts in collaboration with specialists in ethics. We consider penalties that are specified as numbers on a given scale. For illustration purposes, we start by considering a scale from 1 to 3, where a 3-point penalty corresponds to situations with a high gravity. We also account for interactions between penalties and other planning-relevant metrics that should be optimized, such as the total execution time of a plan (as opposed to plan length).

In the remainder of this section, we present the implementation of our framework. We first introduce an adaptation of the \aopl version first presented by Inclezan \citeyear{di23}, which we refer to as \aoplprime. 
We then extend \aoplprime with penalties, resulting in \aoplp (i.e., \aopl with penalties), and expand the corresponding ASP translation. An automated translator from \aoplp into ASP has been implemented and is described next. We then show how reasoning about penalties in planning is handled in our framework, followed by the incorporation of additional metrics, in particular plan execution time. We also revisit behavior modes in this context and discuss a key refinement needed to prevent harm to humans. The section concludes with a high-level overview of the framework.

\subsection{\aoplprime and Its ASP Translation}
\label{sec:aoplprime}

In our work, we build on the \aopl version introduced by Inclezan \citeyear{di23}, which assumes that all rules—including strict ones—are labeled, unlike in the original definition of the language by Gelfond and Lobo (see Section \ref{sec:aopl}). While Gelfond and Lobo assigned labels only to defeasible rules to allow expressing preferences among them, we require labels on both strict and defeasible rules to track which policy rules are violated by an agent. The distinction between defeasible and strict rules specifies only whether a rule can have exceptions to its applicability. It does not determine which rules an agent may violate, as both strict and defeasible applicable rules can be violated.
We further extend this version of \aopl by refining its semantics to support reasoning over trajectories in dynamic systems, a key requirement for planning. 
We refer to this refined version as \aoplprime.

In \aoplprime, rules of a policy may have one of the forms below:
\begin{subequations} \label{eq_prime}
    \begin{align}
        r:    \ \ \ permitted\left(e\right) & \textbf{ if } \ cond \label{eq_prime_1_1}\\[-0.3em]
        r:     \neg permitted\left(e\right) & \textbf{ if } \ cond \label{eq_prime_1_2}\\[-0.3em]
	r:	 \ \ \ \ \ \ \ \ \ \ \ \ \ obl\left(h\right) & \textbf{ if } \ cond \label{eq_prime_1_3}\\[-0.3em]
        r:   \ \ \ \ \ \ \ \ \ \ \   \neg obl\left(h\right) & \textbf{ if } \ cond\label{eq_prime_1_4}\\[-0.3em]
        d: \textbf{normally }  \ \ permitted(e) & \textbf{ if } \ cond \label{eq_prime_2_1}\\[-0.3em]
        d: \textbf{normally }  \neg permitted(e) & \textbf{ if } \ cond \label{eq_prime_2_2}\\[-0.3em]
        d: \textbf{normally }  \ \ \ \ \ \ \ \ \ \ \ \ \ obl(h) & \textbf{ if } \ cond \label{eq_prime_2_3}\\[-0.3em]
        d: \textbf{normally }  \ \ \ \ \ \ \ \ \ \ \ \neg obl(h) & \textbf{ if } 
        \ cond \label{eq_prime_2_4}\\[-0.3em]
	    	 \ \ \textit{prefer}(d_i, d_j) & \label{eq_prime_2_5}
    \end{align}
\end{subequations}
As in the original \aopl language, the rules of \aoplprime use the following notation: $e$ denotes an elementary action, $h$ a happening (i.e., an elementary action or its negation), and $cond$ a set of literals from the signature, excluding those containing the predicate \textit{prefer}.
Rules (\ref{eq_prime_1_1})–(\ref{eq_prime_1_4}) represent strict policy rules, now labeled with a label denoted by $r$.
Defeasible rules of types (\ref{eq_prime_2_1})-(\ref{eq_prime_2_4}) are likewise labeled, as in the earlier version, and contain the keyword \textbf{normally}. 
As before, preference relationships such as rule (\ref{eq_prime_2_5}) can be specified only between defeasible rules, while strict rules always override the defeasible rules with which they conflict.

The semantics of \aoplprime are given in terms of a translation into ASP. We adapt the translation defined by Inclezan \citeyear{di23} in two ways: first, by introducing discrete time steps to reason over trajectories of multiple steps for planning purposes; and second, by modifying the original translation from the \textsc{Clingo} input language to the ASP-Core-2 standard input language for ASP \cite{Calimeri2020}, thereby ensuring compatibility with other solvers such as DLV2 \cite{alviano17}.
 This translation is denoted by  \reilp for an \aoplprime policy \p. 

\medskip
The \textbf{signature} of \reilp for a policy \p applying in a dynamic domain described by a transition diagram \t contains the sorts, statics, fluents, and actions of \t; predicates $action$, $static$, and $fluent$; the sort $step$ representing time steps; a predicate $holds(s)$ for every static $s$; and a predicate $holds(f, i)$ for every fluent $f$ and time step $i$. 

To simplify the presentation of other components of the signature of \reilp, we generalize the syntax of \aoplprime rules of type (\ref{eq_prime_1_1})-(\ref{eq_prime_2_4}) as:
\begin{equation}\label{gen_rule}
r : [{\bf normally}] \ hd \ {\bf if} \  cond
\end{equation}
which refers to both strict and defeasible, authorization and obligation rules from \p. The square brackets ``$[\ ]$'' indicate an optional component of a rule (in this case the keyword ``\textbf{normally}" denoting a defeasible rule).
We use the term  {\em head} of policy rule $r$ to refer to the $hd$ part in (\ref{gen_rule}), where $hd \in HD$,
$$HD = \bigcup\limits_{e \in E}\{permitted(e), \neg permitted(e), obl(e), obl(\neg e), \neg obl(e), \neg obl(\neg e)\}$$ 
and $E$ is the set of all elementary actions in \t. 
We refer to the $cond$ part of a policy rule $r$ as its \textit{body}.
The signature \reilp contains functions $permitted$ and $obl$, as well as a function $neg$ that encodes the negation ``$\neg$'' when it is present in a policy rule. We introduce the following transformation \textbf{lp} that replaces the negation ``$\neg$'' by the function $neg$: 
\begin{itemize}
\item If $x$ is a static, fluent, or elementary action, then \textbf{lp}$(x) = x$ and \textbf{lp}$(\neg x) = neg(x)$

\item If $e$ is an elementary action then:

$
\begin{array}{l}
\textbf{lp}(permitted(e)) = permitted(e)\\
\textbf{lp}(\neg permitted(e)) = neg(permitted(e))\\
\textbf{lp}(obl(e)) = obl(e)\\
\textbf{lp}(\neg obl(e)) = neg(obl(e))\\
\textbf{lp}(obl(\neg e)) = obl(neg(e))\\
\textbf{lp}(\neg obl(\neg e)) = neg(obl(neg(e)))
\end{array}
$
\end{itemize}

The signature of \reilp also includes the following functions:
\begin{itemize}
    \item $b(r)$ for every rule $r$ to denote the condition $cond$ of $r$ (i.e., its body); 
    \item $ab(r)$ for each defeasible rule $r$, representing an exception to its application (i.e., the rule being overridden by another defeasible rule, as specified by a \textit{prefer} relation).

\end{itemize}
\noindent
Additionally, the signature of \reilp contains the predicates:
\begin{itemize}
\item $rule(r)$ -- where $r$ is a rule label (referred shortly as ``rule" below)
\item $type(r, ty)$ -- where $ty \in \{strict,$ \textit{defeasible, prefer}$\}$ is the type of rule $r$
\item $head(r, \textbf{lp}(hd))$ -- to denote the head $hd$ of rule $r$ 
\item $body(r, b(r))$ -- to associate a rule $r$ with its body denoted by function $b(r)$ 
\item $mbr(b(r), \textbf{lp}(l))$ -- for every $l$ in the condition $cond$ of rule $r$, where the condition is represented by $b(r)$
($mbr$ stands for ``member")
\item \textit{prefer}$(d_1, d_2)$ -- where $d_1$ and $d_2$ are defeasible rule labels
\end{itemize}

To reason about which policies are applicable (i.e., active) at each time step, the signature of \reilp also includes the following predicates:
\begin{itemize}
\item $holds(x, i)$ -- where $i$ is a time step and $x$ may be a rule $r$; \textbf{lp}(hd) for the head $hd$ of a rule; the function $b(r)$ representing the body of a rule; or the function $ab(r)$ for every defeasible rule $r$
\item $opp(r, \textbf{lp}(\overline{hd}))$ -- where $r$ is a defeasible rule and  $\overline{hd} \in HD$ ($opp$ stands for ``opposite")
\end{itemize}

The predicate $holds$ determines which policy rules are applicable, based on the truth values of statics and fluents in a state and the interactions among policy rules. Note that $holds$ with arity two is an overloaded predicate, also applying to pairs of fluents and time steps when it is used to describe states of a trajectory in \t.
The predicate $opp(r, \textbf{lp}(\overline{hd}))$ indicates that $\overline{hd}$ is the logical complement of $r$'s head $hd$.

\medskip
The ASP {\bf translation} of a policy \p denoted by the ASP program \reilp consists of:
\begin{enumerate}
    \item A collection \e of facts (or rules) representing the encoding of the policy rules in \p using the predicates $rule$, $type$, $head$, $mbr$, and \textit{prefer}  
    \item The set of policy-independent ASP rules \r shown below in (\ref{eq:independent_rules}), which define predicates $holds(x, i)$ and $opp(r, \textbf{lp}(\overline{hd}))$. In these ASP rules, variable $F$ represents a fluent, $S$ a static, $E$ an elementary action, 
    $H$ a happening (i.e., an elementary action or its negation), and $I$ a time step. 
\begin{equation}
\label{eq:independent_rules}
\begin{array}{lll}
    body(R, b(R)) & \leftarrow & rule(R)\\
    holds(R, I) & \leftarrow & type(R, strict), holds(b(R), I)\\
    holds(R, I) & \leftarrow & type(R, defeasible), holds(b(R), I), \\
             & & opp(R, O), \mbox{not } holds(O, I), \mbox{not } holds(ab(R), I)\\  
    \neg holds (B, I) & \leftarrow & body(R, B), mbr(B, F), fluent(F), \neg holds(F, I).\\
    \neg holds (B, I) & \leftarrow & body(R, B), mbr(B, neg(F)), fluent(F), holds(F, I).\\
    \neg holds (B, I) & \leftarrow & body(R, B), mbr(B, S), static(S), \neg holds(S), step(I).\\
    \neg holds (B, I) & \leftarrow & body(R, B), mbr(B, neg(S)), static(S), holds(S), step(I).\\
    holds(B, I) & \leftarrow &  body(R, B), not \neg holds(B, I), step(I).\\
    holds(ab(R2), I) & \leftarrow & \mbox{\textit{prefer}}(R1, R2), holds(b(R1), I) \\
    holds(Hd, I) & \leftarrow &  holds(R, I), head(R, Hd)\\ 
    opp(R, permitted(E)) & \leftarrow & head(R, neg(permitted(E)))\\
    opp(R, neg(permitted(E))) & \leftarrow & head(R, permitted(E))\\
    opp(R, obl(H)) & \leftarrow & head(R, neg( obl(H)))\\
    opp(R, neg(obl(H))) & \leftarrow & head(R, obl(H))
\end{array}
\end{equation}
\end{enumerate}

Thus, for a policy \p, \reilp = \e $\cup$ \r.

\begin{example}[ASP Encoding \e of a Policy \p]
\label{eg:example1}
Let's give an example of the encoding \e of a policy \p consisting of a unique policy rule:
$$r6(1): obl(stop(1)) \ \textbf{if} \ pedestrians\_are\_crossing(1)$$
where 1 refers to location 1 in Figure \ref{fig:self_driving agent}, $stop(1)$ is an action, and $pedestrians\_are\_crossing(1)$ is a fluent. \e for this policy will consist of the ASP facts:

$
\begin{array}{l}
rule(r6(1)).\\
type(r6(1),\ strict).\\
head(r6(1),\ obl(stop(1))).\\
mbr(b(r6(1)),\ pedestrians\_are\_crossing(1)).
\end{array}
$
\end{example}

\medskip
We discuss next the role of rules \r given in (\ref{eq:independent_rules}). For a given trajectory in the system description to which policy \p belongs, if \p is categorical, atoms of the form $holds(\textbf{lp}(hd), i)$ indicate which literals $hd$ from $HD$ (i.e., literals formed by predicates \textit{permitted} and $obl$) hold at time step $i$. Atoms of the form $holds(r, i)$, where $r$ is a rule and $i$ is a time step, indicate which rules are applicable at each step. If a rule is applicable, the agent’s actions must be checked for compliance. Strict rules apply in all states where their condition (i.e., body) is satisfied, but not in states that do not satisfy it—--see the second rule in (\ref{eq:independent_rules}).
For instance, in the sample policy from Figure \ref{fig:policies}, rule 6 applies only if pedestrians are crossing at the location where the agent is currently situated; crossings at other locations or the absence of pedestrians are irrelevant, and the rule would not apply in those situations. Strict rules with satisfied bodies cannot be rendered inapplicable by rules with the opposite (complement) head. However, a defeasible rule may be overridden by another rule, rendering it inapplicable in a state that would otherwise satisfy its condition—--see the third rule in (\ref{eq:independent_rules}).
The part ``$opp(R, O),\ \mbox{not } holds(O, I)$'' in this ASP rule specifies that an applicable (strict) rule with the opposite head makes the defeasible rule $R$ inapplicable. The part ``$\mbox{not } holds(ab(R), I)$'' indicates that a preference relationship can render a defeasible rule $R$ inapplicable if a more preferred defeasible rule applies in the state. This is encoded as an exception to the default case through the use of $ab(R)$, following the standard treatment of defaults (see the ninth rule in (\ref{eq:independent_rules})). Ultimately, only the policy rules applicable at a given time step $i$ need to be checked for the agent’s compliance at that step.

The reified translation \reilp of \aoplprime into ASP preserves the intended semantics of \aopl. Proposition \ref{prop1} discusses the correspondence between the original ASP translation of \aopl and the reified translation \reilp introduced here for \aoplprime. To establish this correspondence, we first present the ASP translation of a trajectory, which will be referenced in the proposition.
If $t = \langle \sigma_0, ca_0, \sigma_1, ca_1, \dots, ca_n, \sigma_{n+1} \rangle$ is a trajectory in transition diagram \t, where $\sigma_0, \dots, \sigma_{n+1}$ are states and $ca_0, \dots, ca_n$ are actions, then the ASP encoding of $t$ denoted by \et is defined as follows:

$$
\begin{array}{lll}
\mathscr{E}(t) & =_{def} & \{holds(f, i) : f \mbox{ is a fluent},\ f \in \sigma_i,\ 0 \leq i \leq n+1\}\ \cup\\
& & \{\neg holds(f, i) : f \mbox{ is a fluent,}\  \neg f \in \sigma_i\,\ 0 \leq i \leq n+1\}\ \cup\\
& & \{holds(s) : s \mbox{ is a static},\  s \in \sigma_i,\ 0 \leq i \leq n+1\}\ \cup\\
& & \{\neg holds(s) : s \mbox{ is a static},\  \neg s \in \sigma_i,\ 0 \leq i \leq n+1\}\  \cup\\
& & \{occurs(e, i) : e \mbox{ is an elementary action}, \ e \in ca_i, \ 0 \leq i \leq n \}
\end{array}
$$

\begin{proposition}
\label{prop1}
Let \evi be an event in a trajectory $t = \langle \sigma_0, ca_0, \sigma_1, ca_1, \dots, ca_n, \sigma_{n+1} \rangle$ in the transition diagram \t. For every $i \in \{0, \dots, n\}$, let $\mathscr{A}_i$ be the collection of answer sets of $lp(\mathscr{P}, \sigma_i)$ and let  $\mathscr{B}_i$ be the collection of answer sets of \reilp $\cup$ $\mathscr{E}$(\evi).

There is a one-to-one correspondence $map : \mathscr{A}_i \rightarrow \mathscr{B}_i$ such that if $map (A) = B$ then for every $hd \in HD$ where $\textbf{lp}(hd) \in A$, we have that $\exists holds(\textbf{lp}(hd), i) \in B$.
\end{proposition}

\medskip
\textbf{Challenges:} The \reilp translation described above presents specific challenges when it comes to automating the translation process.
The description of \e above assumes that rule labels are ground terms, as in Example \ref{eg:example1}. In practical applications, it is more common for rule labels to contain variables, thus representing a schema for a collection of ground rule labels. If that is the case, then \e would not contain facts, but rather a collection of rules qualifying the variables.

Let's consider rule 6 from Figure \ref{fig:policies} stating that there is a strict obligation to stop when pedestrians are crossing. Assuming that the dynamic domain includes action $stop(l)$ and fluent $pedestrians\_are\_crossing(l)$ where $l$ is a location, the corresponding \aoplprime rule would look as follows:
\begin{equation}\label{eq:rule6.1}
\begin{array}{lll}
r6(L): obl(stop(L)) & \textbf{if} &  pedestrians\_are\_crossing(L)
\end{array}
\end{equation}
Note that the rule label, $r6(L)$ is not ground. Hence the representation of this rule in \e would be as follows:

$
\begin{array}{l}
rule(r6(L)) \ \leftarrow \ action(stop(L))\\
type(r6(L), strict) \ \leftarrow \  rule(r6(L))\\
head(r6(L), obl(stop(L))) \ \leftarrow \  rule(r6(L))\\
mbr(b(r6(L)), pedestrians\_are\_crossing(L)) \ \leftarrow \  rule(r6(L))
\end{array}
$

\noindent
where the body ``$action(stop(L))$ of the first rule above is derived from the name of the action referenced in policy rule 6.

Another challenge posed by the reified translation of \aoplprime into ASP is dealing with arithmetic comparisons. To illustrate this, let's consider rule 1 from Figure~\ref{fig:policies}. The label of the \aoplprime statement for this policy rule, $r1(L_1, L_2, S, S_1)$, needs to keep track of four variables: $L_1, L_2$, and $S$ associated with the $drive$ action from location $L_1$ to location $L_2$ at speed $S$; and the speed limit $S_1$ for the road section from $L_1$ to $L_2$:
\begin{equation}\label{eq:rule1.1}
\begin{array}{lrl}
r1(L_1, L_2, S, S_1) : & \textbf{normally} & \neg permitted(drive(L_1, L_2, S)) \\ & \textbf{if} & speed\_limit(L_1, L_2, S_1),\ S > S_1 + 5, \ S_1 < 55
\end{array}
\end{equation}

The ASP encoding of rule 1 from Figure~\ref{fig:policies} would contain a statement:

$  
\begin{array}{lll}
mbr(b(r1(L_1, L_2, S, S_1)),\ S > S_1+5) & \leftarrow & rule(r1(L_1, L_2, S, S_1))
\end{array}
$

\noindent
Since this syntax is not allowed by ASP solvers, we replace arithmetic comparisons with our own, for example $gt$ for ``$>$'', as in:

$
\begin{array}{lll}
    mbr(b(r1(L_1, L_2, S, S_1)), gt(S, S_1+5)) & \leftarrow &  rule(r1(L_1, L_2, S, S_1))
\end{array}
$

\noindent
and define these new symbols (e.g., $gt, gte$) via ASP rules added to the \r part of \reilp which corresponds to policy-independent ASP rules (see (\ref{eq:independent_rules})).

The process of obtaining these translations is described further in Section \ref{sec:translator}. But first, let's extend \aoplprime with means for representing penalties.

\subsection{Extending \aoplprime with Penalties: \aoplp}
\label{sec:extended_aopl}

In our framework we extend the \aoplprime syntax (i.e., statements of type (\ref{eq_prime_1_1})-(\ref{eq_prime_2_5})) by a new type of statement for penalties:
\begin{equation} \label{eq:penalty_aopl}
    penalty(r, p)\ \textbf{if}\ cond_p
\end{equation}

\noindent 
where \textit{r} is the label of the prohibition or obligation rule for which the penalty is specified, \textit{p} stands for the number of penalty points imposed if the policy rule \textit{r} applies and the agent is non-compliant with it, and $cond_p$ is a collection of static literals. The ``$\textbf{if}\ cond_p$'' part is omitted if $cond_p$ is empty. We denote this extension of \aoplprime as \aoplp.

For instance, the penalty associated with rule 6 from Figure \ref{fig:policies} is encoded in \aoplp as:
\begin{equation}\label{eq:rule6.2}
\begin{array}{l}
penalty(r6(L), 3)
\end{array}
\end{equation}
\noindent
This says that the agent will incur a 3-point penalty at each time step in which this rule applies and the agent’s action is non-compliant with it.

Multiple penalty values may be associated with the same rule, reflecting different gravity levels, as in rule 1 of Figure \ref{fig:policies}. If the rule applies at a given time step and the agent’s action is non-compliant, the agent incurs one of the specified penalties according to the gravity level.
The various levels of penalties assigned to rule 1 
are stated in \aoplp as:
\begin{equation}\label{eq:rule1.2}
\begin{array}{lll}
penalty(r1(L_1, L_2, S, S_1),1) & \textbf{if} & S - S_1 < 10\\
penalty(r1(L_1, L_2, S, S_1),2)  & \textbf{if} &  S - S_1 >= 10,\ 
 S - S1 < 20\\
penalty(r1(L_1, L_2, S, S_1),3)  & \textbf{if} &  S - S_1 >= 20
\end{array}
\end{equation}

\medskip
Recall from the previous section that the semantics of the policy language are given in terms of a translation into ASP. We need to expand this translation to cover the new type of statement in (\ref{eq:penalty_aopl}). 
Given a policy \p, recall that \reilp = \e $\cup$ \r, where \e is the encoding of the statements in \p and \r is the set of policy-independent rules described in (\ref{eq:independent_rules}). We expand the \e part of \reilp with the ASP translation of the new type of \aoplp statement (\ref{eq:penalty_aopl}) for penalties:
$$
\begin{array}{lll}
penalty(r, p) & \leftarrow & rule(r),\ cond\_p
\end{array}
$$

\noindent
The rules in \r remain as they are.
As an example, statement (\ref{eq:rule6.2}) is translated into ASP as:

$
\begin{array}{lll}
penalty(r6(L), 3) & \leftarrow & rule(r6(L))
\end{array}
$

\noindent
Penalty statements (\ref{eq:rule1.2}) are translated as:

$
\begin{array}{l}
penalty(r1(L_1, L_2, S, S_1),1) \ \leftarrow \ rule(r1(L_1, L_2, S, S_1)),\ S - S_1 < 10\\
penalty(r1(L_1, L_2, S, S_1),2)  \ \leftarrow \  rule(r1(L_1, L_2, S, S_1)),\ S - S_1 \geq 10,\ S - S_1 < 20\\
penalty(r1(L_1, L_2, S, S_1),3)  \ \leftarrow \  rule(r1(L_1, L_2, S, S_1)),\ S - S_1 \geq 20
\end{array}
$

\medskip
We developed a Python-based translator from the  \aoplp into ASP that creates the ASP policy encoding \e for a policy \p. We describe this translator next. 

\subsection{Translator from \aoplp to ASP}
\label{sec:translator}
To automate the translation of \aoplp policies and their penalties and streamline the use of our ASP-based framework, we developed the first Python-based translator for \aoplp. This translator takes as input a text file containing all relevant \aoplp policy rules for a given domain, including associated penalties, and generates the corresponding ASP encodings for all policy rules and penalties as described in Sections \ref{sec:aoplprime} and \ref{sec:extended_aopl}. 
This automated translation produces ASP rules defining predicates $rule$, $type$, $head$,  $mbr$, \textit{prefer}, and $penalty$ and corresponds to the policy-specific component \e of the program \reilp for a policy \p. 

For example, consider the \aoplp representation of policy rule 1 from Figure~\ref{fig:policies} shown in (\ref{eq:rule1.1}) and its associated penalties seen in (\ref{eq:rule1.2}), copied below for clarity:
$$
\begin{array}{lrl}
r1(L_1, L_2, S, S_1) : & \textbf{normally} & \neg permitted(drive(L_1, L_2, S)) \\ & \textbf{if} & speed\_limit(L_1, L_2, S_1),\ S > S_1 + 5, \ S_1 < 55
\end{array}
$$
$$
\begin{array}{lll}
penalty(r1(L_1, L_2, S, S_1),1) & \textbf{if} & S - S_1 < 10\\
penalty(r1(L_1, L_2, S, S_1),2)  & \textbf{if} &  S - S_1 >= 10,\ 
 S - S1 < 20\\
penalty(r1(L_1, L_2, S, S_1),3)  & \textbf{if} &  S - S_1 >= 20
\end{array}
$$
According to policy rule 10 in Figure \ref{fig:policies}, $r1$ overrides $r7$, which corresponds to the \aoplp statement:
$$
\textit{prefer}(r1(L_1, L_2, S, S_1), r7(L_1, L_2, S))
$$
The expected and generated output of the translator for rule $r1$, as well as its associated penalties and \emph{prefer} statement, is shown in Figure \ref{fig:listing}.

\begin{figure}[hbt!]
    \centering
\begin{lstlisting}
rule(r1(L1, L2, S, S1)) :-
    action(drive(L1, L2, S)),
    holds(speed_limit(L1, L2, S1)).
type(r1(L1, L2, S, S1), defeasible) :- rule(r1(L1, L2, S, S1)).
head(r1(L1, L2, S, S1), neg(permitted(drive(L1, L2, S)))) :- 
    rule(r1(L1, L2, S, S1)).
mbr(b(r1(L1, L2, S, S1)), speed_limit(L1, L2, S1)) :- 
    rule(r1(L1, L2, S, S1)).
mbr(b(r1(L1, L2, S, S1)), gt(S, S1+5)) :- rule(r1(L1, L2, S, S1)).
mbr(b(r1(L1, L2, S, S1)), lt(S1, 55)) :- rule(r1(L1, L2, S, S1)).
prefer(r1(L1, L2, S, S1), r7(L1, L2, S)) :- 
    rule(r1(L1, L2, S, S1)), rule(r7(L1, L2, S)).
penalty(r1(L1, L2, S, S1),1) :- rule(r1(L1, L2, S, S1)), S - S1 < 10.
penalty(r1(L1, L2, S, S1),2) :- rule(r1(L1, L2, S, S1)), S - S1 >= 10, 
    S - S1 < 20.
penalty(r1(L1, L2, S, S1),3) :- rule(r1(L1, L2, S, S1)), S - S1 >= 20.
\end{lstlisting}
\caption{ASP translation for policy rule 1 from Figure \ref{fig:policies}}
    \label{fig:listing}
\end{figure}

The rule $r1$ highlights several challenges for the translator in generating safe rules for a solver like \textsc{Clingo}.   
First, to ensure the safety of the policy rule definition (see lines 1–3 in the listing in Figure \ref{fig:listing}), the translator must not only identify the action being permitted or prohibited---in this case, $drive(L_1, L_2, S)$---but also account for any other variables in the rule label not directly associated with the action. These variables must be linked to an appropriate fluent or static predicate and added to the body of the ASP rule. In this case, $S_1$ is unrelated to the $drive(L_1, L_2, S)$ action mentioned in the head of the \aoplp rule (i.e., before the keyword \textbf{if}), and instead represents the speed limit on the road segment from $L_1$ to $L_2$. The translator identifies $speed\_limit(L_1, L_2, S_1)$ in the body of the \aoplp rule as the relevant static and includes $holds(speed\_limit(L_1, L_2, S_1))$ in the body of the corresponding ASP rule (see line 3 in the listing). 
A second issue is that the translator must determine whether a policy rule is defeasible or strict based on the presence or absence of the keyword \textbf{normally}, adding a corresponding rule to the translation that specifies its type as either $type(r, defeasible)$ or $type(r, strict)$ (see line 4 in the listing).
Finally, the body of the policy rule contains comparison operators, which must be translated into predicates such as $gt$ or $lt$ in the reified logic programming encoding. These predicates ensure that comparisons can be properly represented as terms within the $mbr$ predicate (see lines 9–10 in the listing).

Our Python translator relies on the following functions to achieve this:
\begin{itemize}
\item $parse\_aopl\_rule()$: decomposes an \aoplp policy into its fundamental components; extracts rule labels; separates the rule into head (authorization or obligation for an action) and body (conditions); and analyzes variables for classification and further processing.

\item $classify\_literals()$: processes the body of the rule to format it for ASP; filters out numerical conditions (e.g., inequalities); wraps state-related conditions in $holds$ predicates; and classifies literals based on whether they pertain to actions or states.

\item $translate\_aopl\_rule\_to\_asp()$: converts parsed \aoplp rules into ASP syntax; constructs the rule $head$ using $permitted$ and $obl$ constructs and assembles the body $mbr$ by incorporating conditions and constraints.

\item $aopl\_to\_asp()$: processes different types of \aoplp rules, determining whether they are strict or defeasible; translates penalty clauses and preference statements into the corresponding ASP representation.

\item $process\_aopl\_file()$: manages file input and output for the translator; reads \aoplp policies from a text input file, processes them using $aopl\_to\_asp()$, and writes the translated ASP rules to an output text file.
\end{itemize}

The code for the translator is available at \url{https://github.com/vineelsai313/Penalties}.

\subsection{Reasoning about Penalties in Planning}

In this section we present an ASP reasoning mechanism that  considers penalties in planning. We have shown in Section \ref{sec:aoplprime} how the program \reilp can be used to determine which policies are applicable (i.e., are active and hence should be checked for compliance with) at each step in a trajectory. The policy rules applicable at each step may be strict or defeasible. An agent can choose to be non-compliant with either type; the distinction simply indicates whether a rule allows exceptions to its applicability.

We start by devising ASP rules that can flag what penalty is incurred by an agent at a time step, and for being non-compliant with which policy rule. We do so by introducing a predicate $add\_penalty(r, p, i)$, which says that a penalty of $p$ points should be incurred for non-compliance with an applicable policy rule $r$ at time step $i$. The definition of this predicate covers three cases for non-compliance. The first one is when an action that is not permitted to be executed is included in the plan:

$
\begin{array}{lll}
 add\_penalty(R, P, I) & \leftarrow & rule(R), \ 
holds(R, I), \ head(R, neg(permitted(E))), \\
 & & occurs(E, I),\ penalty(R, P)
\end{array}
$

\noindent
where \textit{R} is a rule, \textit{I} is a time step, \textit{E} is an elementary agent action, and \textit{P} is a  penalty. 

The second case is when an agent is obligated to execute an action but the action is not included in the plan:

$
\begin{array}{lll}
 add\_penalty(R, P, I) & \leftarrow & rule(R), \ 
holds(R, I), \ 
 head(R, obl(E)), \\ 
 & & action(E), \mbox{not } occurs(E, I), \ penalty(R, P)
\end{array}
$

The final case is when an agent is obligated to not execute an action, but the action is included in the plan:

$
\begin{array}{lll}
 add\_penalty(R, P, I) & \leftarrow & rule(R), \ 
 holds(R, I), \ 
 head(R, obl(neg(E))), \\
 & & occurs(E, I), \ penalty(R, P)
\end{array}
$

The overall penalty for a plan can be captured by the predicate $cumulative\_penalty$ defined using the $\#sum$ aggregate:

$
\begin{array}{lll}
    cumulative\_penalty(N) & \leftarrow &  \#sum\{P, R, I: add\_penalty(R, P, I)\} = N
\end{array}
$

Introducing this predicate is optional as this information can be retrieved from the solver's output, but we include it here to make the overall penalty of plans clearer in later examples.

\begin{example}
\label{ex:example1}
Consider the Traffic Norms domain from Section \ref{sec:example} and an agent that starts its journey at location 6 with the objective of reaching destination location 10. A critical point of this route is that there is a ``Do not enter'' sign situated at location 8, and a school bus is stopped between 14 and 13 at time step 3. Let's analyze a possible plan and the associated penalties:
\begin{equation*}
    \begin{array}{lll}
\text{occurs(drive(6,8,45),0)} & \ \ \ \ \ & \text{add\_penalty(r3(6,8,45),3,0)} \\
\text{occurs(drive(8,7,85),1)} & & \text{add\_penalty(r1(6,8,45,15),3,0)}\\
\text{occurs(drive(7,9,45),2)} & & \text{add\_penalty(r1(8,7,85,45),3,1)}\\
\text{occurs(drive(9,10,85),3)} & & \text{add\_penalty(r1(7,9,45,15),3,2)}\\
\text{cumulative\_penalty(15)} &  & \text{add\_penalty(r1(9,10,85,25),3,3)} \\
\end{array}
\end{equation*}
Recall that the third parameter of a $drive$ action is the speed in mph. Looking at the incurred penalties, note that there are two penalties of 3 points added at time step 0, one for breaking rule 3 by disobeying the ``Do not enter'' sign at location 8, and another one for rule 1 for speeding between locations 6 and 8 with a speed of 45 mph while the speed limit on that section is set to 15 mph.
\end{example}

\subsection{Adding Other Metrics: Time}
One of our goals is to introduce a distinction between different Risky behavior plans (those that allow non-compliant actions). So far we considered cumulative penalty as one of the metrics. Another metric, which corresponds to reasoning about emergency situations, is time (i.e., duration). This is in fact a more accurate metric than the one used in previous work for emergency situations---plan length---and more generalizable to other scenarios, as compared to other metrics such as distance. A new predicate is added to our ASP framework, $add\_time(t, i)$, which means add $t$ time units to the overall plan execution time, 
to account for the duration of the action executed at time step $i$. Rules that define this predicate look like:

$
\begin{array}{lll}
    add\_time(t, I) & \leftarrow & occurs(e, I)
\end{array}
$

\noindent
where $t$ is a number representing the time units and $e$ is an elementary action.

To test our framework, we assigned a certain number of time units for each action, as follows: 5-time units if driving between two connected locations at a speed $>$ 55 mph; 10-time units if driving at a speed between 36 mph and 55 mph; 15-time units if driving at a speed that is $\leq$ 35 mph; and 2-time units  if stopped. Here is an example in ASP:

$
\begin{array}{lll}
add\_time(5, I)  & \leftarrow & occurs(drive(L_1, L_2, S), I), \ S > 55
\end{array}
$

Overall time to execute a plan can be calculated similarly to cumulative penalty, by introducing a $cumulative\_time$ predicate:

$
\begin{array}{lll}
    cumulative\_time(N) & \leftarrow & \#sum\{T,I: add\_time(T, I)\} = N
\end{array}
$

\begin{example}
Consider the scenario in Example \ref{ex:example1}. The following $add\_time$ and $cumulative\_penalty$ atoms would characterize the plan in that example:

$
    \begin{array}{lllll}
\text{add\_time(10,0)} & \ \ \ \ \ & \text{add\_time(10,2)} & \ \ \ \ \ & \text{cumulative\_time(30)}\\
\text{add\_time(5,1)} & \ \ \ &  \text{add\_time(5,3)} & & \\
\end{array}
$
\end{example}

\subsection{Behavior Modes Revisited}
\label{sec:behavior_modes}
Different behavior modes can be expressed as priorities between the metrics discussed above, cumulative penalty and cumulative time, as well as other potential ones. As a first distinction, we introduce the \textbf{emergency} and \textbf{non-emergency} behavior modes. The former prioritizes time over penalties, while the latter does the opposite. The behavior mode is set by the agent's controller by adding one of the two facts below:

$
\begin{array}{lll}
emergency & \mbox{   \textbf{or}   } & non\_emergency
\end{array}
$

\noindent
Priorities for time versus penalty for each of these modes are set via the rules:

$
\begin{array}{lll}
    time\_priority(2) & \leftarrow &  emergency\\
    penalty\_priority(1) & \leftarrow &  emergency\\
    penalty\_priority(2) & \leftarrow & non\_emergency\\
    time\_priority(1) & \leftarrow &  non\_emergency\\
\end{array}
$

$
\begin{array}{l}
:\sim add\_penalty(R, P, I), penalty\_priority(Y).\ [P@Y, R, I]\\
:\sim add\_time(T, I), time\_priority(Y).\ [T@Y, I]		  
\end{array}
$

\noindent
This encoding relies on the fact that the larger the number following the ``$@$'' symbol in a soft constraint, the higher the priority of complying with that constraint.

\begin{example}
For the scenario in Example \ref{ex:example1}, an agent in the \textbf{emergency} mode would choose the plan:

$
\begin{array}{lll}
occurs(drive(6,8,45),0) & occurs(drive(8,7,85),1) & occurs(drive(7,9,45),2)\\
occurs(drive(9,10,85),3) & & 
\end{array}
$

\noindent
with cumulative penalty 15 and cumulative time 30. An agent in the \textbf{non-emergency} mode would choose the plan: 

\noindent
$
\begin{array}{lll}
occurs(drive(6,11,15),0) & occurs(drive(11,12,65),1) &   occurs(drive(12,14,15),2)\\
  occurs(stop(14),3) &  occurs(drive(14,13,25),4) & occurs(drive(13,10,15),5)
\end{array}
$

\noindent
with cumulative penalty 0 and cumulative time 67. This agent would stop at location 14 at time step 3 as there is a stopped school bus between 13 and 14.
\end{example}

If the \aoplp policy is both consistent and unambiguous, then the ASP program for each of the behavior modes {\bf emergency} vs {\bf non-emergency} is guaranteed to compute the optimal plan with respect to that behavior mode. 

\subsection{Discussion}
\label{sec:discussion}

As mentioned in Section \ref{sec:example}, penalties are set initially on a scale from 1 to 3 depending on the severity of the infraction. However, not harming humans should be of utmost importance. Therefore, in the Traffic Norms domain, an agent should not violate policy rules that involve pedestrians and school buses. One way to achieve this is to assign a high penalty for non-complying with those policy rules (e.g., a penalty of 50 points, but this number can be changed according to the domain), and flag this value via a predicate $high\_penalty$. To prevent the harm of human life, we can then add a constraint:
$$\leftarrow \ add\_penalty(R, H, I),\  high\_penalty(H)$$
This ensures that the agent always complies with applicable policy rules whenever non-compliance could result in human harm.

To account for situations like the ones in the thought experiment called ``The Trolley Problem'' developed by philosopher Philippa Foot and adapted by Judith Jarvis Thomson, such a constraint may be excluded by the controllers of the agent or policy makers, and replaced with a directive to minimize the execution of actions that result in human harm (i.e., minimize actions that incur a high penalty, represented as a soft constraint):

$
:\sim add\_penalty(R, H, I),\ high\_penalty(H).\ [H@3, R, I]
$

\noindent
Variations of the rules above can be devised for different scenarios. 

Furthermore, other behavior modes can be specified, for instance one in which the agent looks for plans with an upper bound  for the overall time  and (otherwise) minimum penalty (which corresponds to situations in real life where someone needs to arrive to the destination within a time limit, but tries to do so with minimum penalty):

$
\begin{array}{l}
    \leftarrow \ culumative\_time(N),\  max\_time(M),\ N > M\\
    :\sim add\_penalty(R, P, I), penalty\_priority(Y).\ [P@Y, R, I]
\end{array}
$

\subsection{High-Level View of the Framework}

\begin{figure} [htb]
    \centering
    \includegraphics[width = 0.9\textwidth]{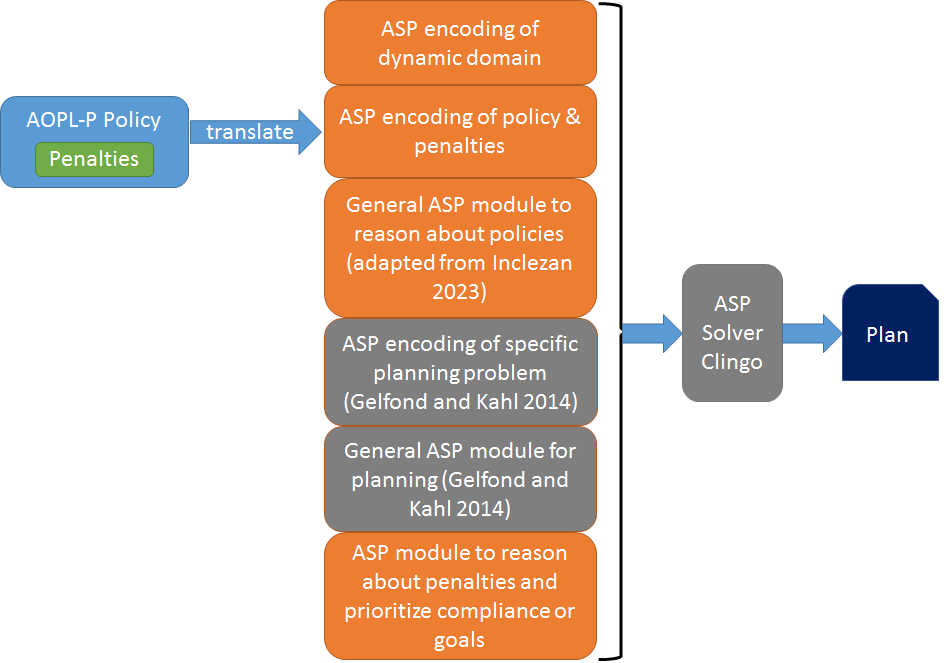}
    \caption{High-Level Framework View}
    \label{fig:framework}
\end{figure}

Key components of our ASP framework are depicted in Figure \ref{fig:framework}. Grey elements indicate modules that are adopted from work by others, and non-grey components are either our own or adapted from others' work with substantial modifications.
The main components of the framework are:
\begin{enumerate}  
    \item The ASP encoding of the \textbf{dynamic domain}, written according to established ASP methodologies (e.g., \cite{gk14}).

    \item The ASP encoding of the \textbf{policy and its associated penalties}. Policies for the dynamic domain are initially specified in \aoplp, together with their penalties. Then \aoplp statements are automatically translated into ASP using the translator described in Section \ref{sec:translator}. This corresponds to the \e component of the program \reilp for a policy \p.
    
    \item A \textbf{general ASP module for reasoning about policies.} This module determines which policies are applicable at different time steps in the execution of a plan. It corresponds to the policy-independent component \r of the program \reilp, substantially adapted from work by Inclezan \citeyear{di23}.

    \item An ASP encoding of a \textbf{specific planning problem}, specifying the initial state, goal state, any observations along the way, and whether the situation is an emergency or non-emergency, following established methods for ASP planning \cite{gk14}.
    
    \item A \textbf{general ASP module for planning} developed according to established ASP planning methodologies, to generate plans \cite{gk14,SonPBS23}.
    
    \item An ASP module to \textbf{rank plans and select the optimal plan}. This module captures the decision making process of a policy-aware agent in emergency versus non-emergency scenarios, by prioritizing actions based on a combination of penalty severity and time-efficient goal achievement. 
\end{enumerate}

In our experimental setup, these ASP components are combined and provided to an ASP solver to compute the optimal plan.

\section{Experimental Results}
\label{sec:experimental_resuts}
Our framework is meant to refine agent behavior specification and simulation---as compared to previous work by Harders and Inclezan \citeyear{hi23}, which we call here the \textit{HI framework}---for situations when non-compliant actions need to be included in a plan due to a high-stakes goal. By introducing penalties and additional metrics such as plan execution time, we improve on the quality of plans generated by the HI framework, since minimizing plan length does not necessarily result in minimal execution time. We also account for minimizing harm to humans, which the HI framework does not address. We report the relative time performance and plan quality of the two frameworks in Table~\ref{tbl:performance_room} for the \textit{Rooms Domain} \cite{hi23}, and in Table~\ref{tbl:performance_traffic} for the newly introduced \textit{Traffic Norms Domain} described in Section~\ref{sec:example}.  
All the tests were performed on a computer with a 12th Gen Intel(R) Core(TM) i5-12500H 2.50 GHz CPU and RAM 16GB. We used the \textsc{Clingo} solver, as in the HI Framework by Harders and Inclezan (to facilitate comparisons), and leveraged some of its specific constructs (e.g., $\#minimize$ in place of soft constraints). 
The code used in these experiments can be found at \url{https://github.com/vineelsai313/Penalties}.

\begin{figure} [htb]
    \centering
    \includegraphics[width = 0.6\textwidth]{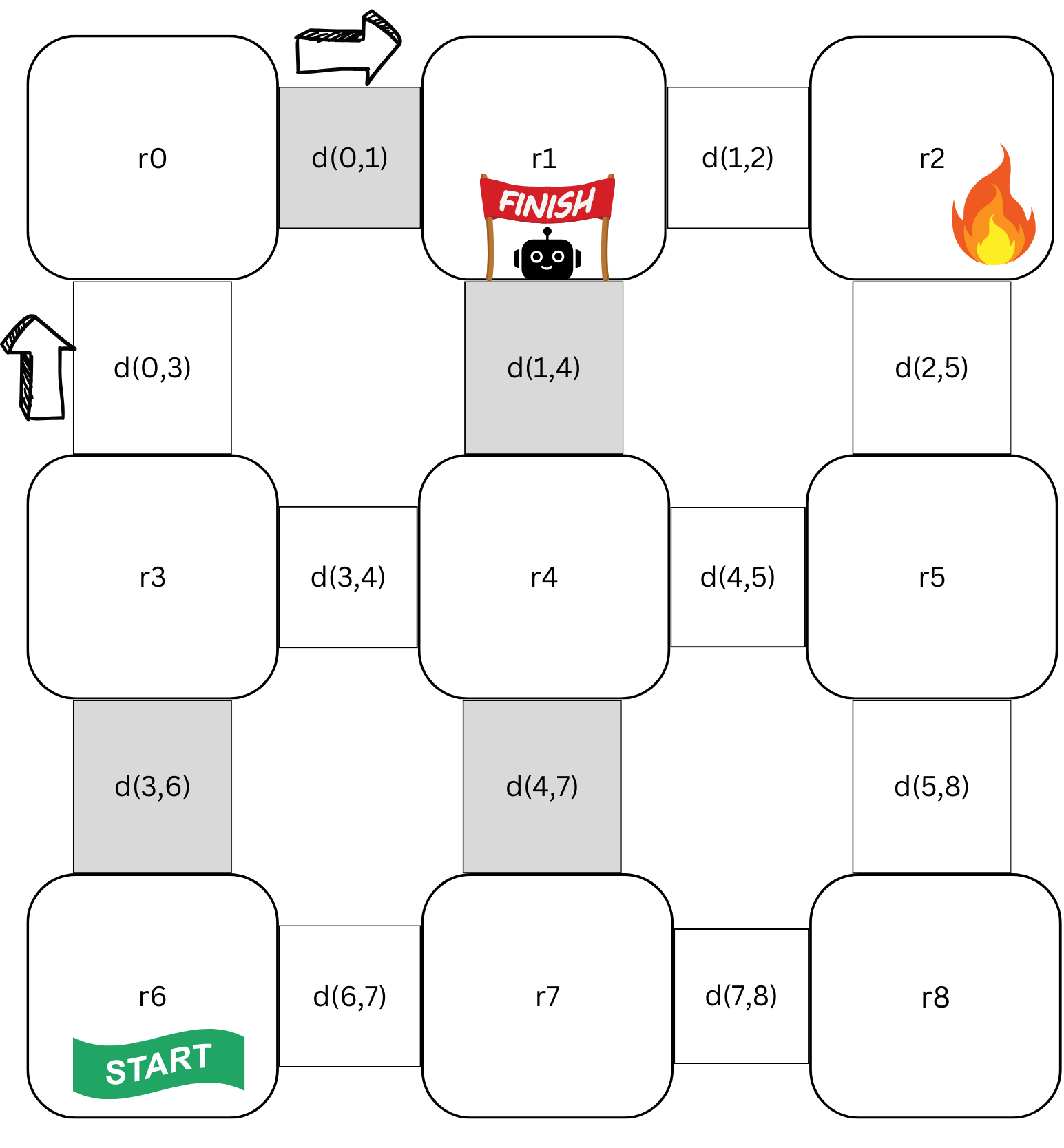}
    \caption{Rooms Domain -- Scenario \#3. The agent starts in room $r6$ and needs to get to room $r1$. Arrows indicate uni-directional doors. White doors are unlocked, while grey doors are locked. There is an active fire in room $r2$.}
    \label{fig:room_domain}
\end{figure}

The Rooms Domain consists of nine rooms labeled from $r0$ to $r8$, as shown in Figure \ref{fig:room_domain} for a sample scenario. Rooms may be connected by doors, some of which may be one-way doors. Doors can be locked and unlocked by the agent using either a key specific to a door or a badge that can open any door. 
The agent starts in one of the rooms and aims to reach a specified destination. Additionally, the agent has knowledge of extreme conditions, such as active fires or contamination in certain rooms, and may possess special protective equipment.  

This domain is governed by both authorization and obligation policies, which include both strict and defeasible rules. Some of these policies are:  
\begin{enumerate}
    \item The agent is obligated to use the key before using the badge if it possesses both.
    \item The agent is not permitted to use its badge more than three times. 
    \item The agent is not permitted to open a one-way door from the wrong side.
    \item Normally, the agent is obligated not to enter a room where there is an active fire.
    \item However, if the agent has special protective equipment, it is allowed to enter a room with an active fire.
    \item Normally, the agent is not permitted to enter a contaminated room.
\end{enumerate}
We further refined these policies by introducing penalty points ranging from 1 to 3, assigned based on the severity of the action.

We present the results for the Rooms Domain in Table~\ref{tbl:performance_room}. Our non-emergency mode corresponds to the Normal mode in the HI framework, which prioritizes plan length while disallowing non-compliant actions (see Section \ref{sec:HI_framework}). Meanwhile, our emergency mode aligns with the Risky mode, which disregards policies entirely -- note the similar plan lengths between these modes. As the table indicates, our penalization framework shows significant improvements in time efficiency, with the experiments for our framework completing in under 0.5 seconds, while the HI framework takes 2--4 seconds to complete each experiment on the same machine. We believe this is because the HI framework employs more metrics that are more complex (e.g., percentage of strongly-compliant elementary actions in a plan) than the ones in our framework.

\medskip
The performance results for the Traffic Norms Domain in Table \ref{tbl:performance_traffic} show higher runtimes. This is because generating plans in this domain does not involve only selecting the path that the agent must follow, but also choosing a driving speed for each road segment, out of a set of available speeds. We noticed increased runtimes for higher cardinalities of the set of possible speeds to select from (Scenarios \#1-10 had six driving speed values to choose from; Scenarios \#11-13 had eleven driving speed values). 
Other causes of an increased runtime are the additional obstacles that the agent must navigate in each scenario compared to earlier ones. 
\small
\begin{table}[hbt!]
\caption{Performance Results: Room Domain}
\label{tbl:performance_room}
\centering
\begin{tabular*}{\linewidth}{@{\extracolsep{\fill}} c  r r  r r  r r  r r }
\hline\hline
         & \multicolumn{4}{c}{\textbf{Our Proposed Framework}}  & \multicolumn{4}{c}{\textbf{The HI Framework}} \\ \hline
         & \multicolumn{2}{c}{\textbf{Non-Emergency}}  & \multicolumn{2}{c}{\textbf{Emergency}}  & \multicolumn{2}{c}{\textbf{Normal Mode}}  & \multicolumn{2}{c}{\textbf{Risky Mode}} \\ 
Scenario \# & T (s)     & L & T (s) & L & T (s) & L & T (s) & L \\ \hline
1        & 0.366        & 3           & 0.267    & 3    & 2.989       & 3           & 3.078    & 3       \\ 
2        & 0.254        & 5           & 0.225     & 4   & 3.267       & 5           & 3.086    & 4          \\ 
3        & 0.227        & 6           & 0.202    & 5    & 2.831       & 6           & 2.780    & 5        \\ 
4        & 0.286         & 10          & 0.203    & 3   & 3.325       & 10          & 3.360     & 3          \\ 
5        & 0.239         & 7           & 0.411    & 2   & 3.014       & 7           & 2.267    & 2         \\ 
6        & 0.285        & 3           & 0.277    & 3    & 2.717       & 3           & 2.969    & 3          \\ 
7        & 0.253        & 5           & 0.328    & 3    & 3.158       & 5           & 2.555    & 3         \\ 
8        & 0.297         & 4           & 0.421    & 4   & 2.559       & 4           & 3.180    & 4          \\ 
9        & 0.325        & 3           & 0.247     & 3   & 2.897       & 3           & 2.319    & 3         \\ 
10       & 0.178        & 0           & 0.150    & 0    & 3.116       & 0           & 3.034    & 0       \\ 
11       & 0.230        & 4           & 0.238    & 3    & 3.399        & 4           & 2.766    & 3        \\ 
12       & 0.274        & 3           & 0.227    & 3    & 3.074       & 3           & 3.050    & 3         \\ 
13       & 0.270        & 6           & 0.389     & 6   & 2.959        & 6           & 2.636    & 6        \\ 
14       & 0.281        & 4           & 0.239     & 4   & 2.616       & 4           & 2.836    & 4          \\ \hline \hline
\end{tabular*}
T (s) -- average time in seconds over 10 runs; L -- plan length\\
\end{table}
\begin{table}[hbt!]
\centering
\caption{Performance Results: Traffic Norms Domain}
\label{tbl:performance_traffic}
\begin{tabular*}{\linewidth}{@{\extracolsep{\fill}} c  r r  r r r r r r}
\hline\hline
         & \multicolumn{4}{c}{\textbf{Our Proposed Framework}}  & \multicolumn{4}{c}{\textbf{The HI Framework}} \\ \hline
         & \multicolumn{2}{c}{\textbf{Non-Emergency}}  & \multicolumn{2}{c}{\textbf{Emergency}}  & \multicolumn{2}{c}{\textbf{Normal Mode}}  & \multicolumn{2}{c}{\textbf{Risky Mode}} \\ 
Scenario \# & T (s)     & L & T (s) & L & T (s) & L & T (s) & L \\ \hline
1        & \ \ \ 3.853        & 6           & \ \ \ 4.108    & 4  &1.840  &6  &1.559  &4      \\ 
2        & \ \ \ 4.494        & 2           & \ \ \ 3.678    & 2  &1.731  &2  &1.134  &2     \\
3        & \ \ \ 3.902        & 4           & \ \ \ 3.944    & 4  &0.738  &4  &0.739  &4      \\ 
4        & \ \ \ 4.252        & 5          & \ \ \ 4.682     & 5  &0.722  &5  &0.598  &3      \\ 
5        & \ \ \ 3.644        & 3           & \ \ \ 3.599    & 3  &0.715  &3  &1.808  &2      \\ 
6        & \ \ \ 4.660        & 4           & \ \ \ 3.661    & 4  &1.631  &4  &0.773  &3      \\
7        & \ \ \ 4.852        & 3           & \ \ \ 4.837    & 3  &0.592  &3  &1.001 &3      \\
8        & \ \ \ 3.624         & 4           & \ \ \ 5.005   & 4  &0.759  &4  &0.603  &3      \\    
9        & \ \ \ 3.634          & 3           & \ \ \ 5.510  & 3  &0.887  &3  &0.681  &3      \\
10       & \ \ \ 5.283        & 4           & \ \ \ 3.731    & 3  &0.649  &4  &0.618 &3      \\
11       & \ \ \ 18.627       & 6           & \ \ \ 12.558   & 6  &0.763  &6  &1.198  &5      \\  
12       & \ \ \ 17.492       & 6           & \ \ \ 15.853   & 6  &0.869  &6  &0.602  &5      \\
13       & \ \ \ 17.085       & 6           & \ \ \ 18.223   & 6  &0.686  &9  &0.940  &6      \\
\hline \hline
\end{tabular*}
T (s) -- average time in seconds over 10 runs; L -- plan length
\end{table}
\normalsize
The results in Table \ref{tbl:performance_traffic} indicate that the HI framework is more efficient than ours on the Traffic Norms Domain, but this is in the detriment of plan quality. Due to the nature of the Traffic Norms Domain in which selecting the appropriate speed is important, especially when penalties are considered, our framework goes through substantially more optimization cycles (3-16 vs 2-6 for the HI framework). This is a main cause for the performance differences for this domain. The HI framework simply finds a plan that it deems optimal sooner, because it does not work as hard on refining  driving speeds.

In terms of plan quality, the plans produced by the HI Framework in the Risky mode are sometimes shorter than ours in the emergency mode (see Scenarios \#4-6, 8, and 11-12). This is because our agent stops in situations when humans are involved (pedestrians or stopped school buses as described in Section \ref{sec:discussion}) even in the emergency mode, while the HI Risky agent does not and may lead to harming humans. In most occasions, however, the plans produced by the two frameworks do not differ with respect to the path chosen by the agent but they may differ in the driving speed that is selected. For instance, a Normal agent's plan under the HI framework may include multiple actions of driving substantially under the speed limit, e.g., at 5 mph on road segments where the speed limit is 25 mph. In real life, such situations are undesirable as they may prompt other drivers to unsafe road behavior. 
In contrast, an agent in our non-emergency mode would choose driving speeds that follow closely the speed limit because total time to destination is considered in the selection of an optimal plan, though with less of a priority in the non-emergency mode, and different driving speeds imply different time needed to complete the action. 
In the Risky mode, the HI framework may select speeds that are either very low as before, or very high compared to the speed limit, because it deems all non-compliant actions equal. In our framework, the higher the driving speed is over the speed limit, the higher the penalty that is paid. Moreover, if an action breaks two policy rules, under the HI framework this would be counted as one non-compliant action, while in our framework, the agent pays the corresponding penalties for each of the two infractions. 

\begin{example}[Sample Plan Comparison: Our Framework vs. the HI Framework]
To highlight the difference between the plans generated by our framework and those from the HI framework, consider Scenario \#5 from the Traffic Norms Domain—specifically in emergency mode, which corresponds to the Risky behavior mode in the HI framework. In this scenario, the agent starts at location 6 at time step 0 and aims to reach location 4. The speed limit between locations 6 and 5 is 25 mph, and between 5 and 4 it is 15 mph, with similar speed limits on other segments. Additionally, pedestrians are crossing at location 5 at time step 1.  
As shown in Table \ref{tbl:performance_traffic}, our framework produces a plan of length 3, whereas the HI framework generates a shorter plan of length 2. The plan generated by our framework is as follows:

$
\begin{array}{lll}
occurs(drive(6,5,85), 0)  &
occurs(stop(5), 1) &
occurs(drive(5,4,45), 2)
\end{array}
$

\noindent
The HI Framework generates the following plan:

$
\begin{array}{ll}
occurs(drive(6,5,65), 0) &
occurs(drive(5, 4, 15), 1)
\end{array}
$

\noindent
Notably, even in emergency mode, our framework's plan includes stopping for crossing pedestrians at time step 1, reflecting its design to prevent harm to humans even under high-stakes conditions. In contrast, the agent guided by the HI framework proceeds without stopping at time step 1. Additionally, the HI framework does not optimize driving speeds to minimize total time, as it considers only plan length, regardless of the urgency or context of the situation.
\end{example}

In the Traffic Norms Domain, we examined how the number of distinct speed values affects performance within our framework. We ran scenarios 1–10 from Table \ref{tbl:performance_traffic} using the original six speed values, as well as configurations with seven and eight speed values. The results are presented in Table \ref{tbl:performance_traffic-multiple_speeds}. As expected, the average computation time increases as the number of considered speed values grows. Some scenarios (e.g., scenario \#5) are less affected, likely because the origin and destination are closer on the map, resulting in inherently shorter plans. The non-emergency mode is more sensitive to increased number of different speeds due to the stricter enforcement of policy rules in this mode, many of which directly involve speed constraints.

\begin{table}[hbt!]
\caption{Impact of the Number of Distinct Speed Values on Performance in the Traffic Domain}
\label{tbl:performance_traffic-multiple_speeds}
\centering
\begin{tabular*}{\linewidth}{@{\extracolsep{\fill}} c  c c  r r  r r  r r r r }
\hline\hline
\# of Speeds & \multicolumn{2}{c}{\textbf{6 Speed Values}}  & \multicolumn{2}{c}{\textbf{7 Speed Values}} & \multicolumn{2}{c}{\textbf{8 Speed Values}}\\ \hline
 & \textbf{Non-Em.}   & \textbf{Em.}  & \textbf{Non-Em.} & \textbf{Em.}  & \textbf{Non-Em.} & \textbf{Em.} \\ 
Scenario \# & T (s)     & T (s) & T (s) & T (s) & T (s) & T (s)\\ \hline
1        & 3.853 & 4.108  &   5.626 &  5.907 &   9.986 &  10.024\\ 
2        & 4.494 & 3.678  &   3.653 &  3.989 &   8.936 &  8.731\\ 
3        & 3.902 & 3.944  &   4.457 &  6.853 &   7.839 &  8.867\\ 
4        & 4.252 & 4.682  &   6.282 &  8.656 &   8.264 &  12.325\\ 
5        & 3.644 & 3.599  &   4.845 &  5.436 &   3.911 &  4.903\\ 
6        & 4.660 & 3.661  &   5.901 &  3.723 &   11.372 &  4.813\\ 
7        & 4.852 & 4.837  &   4.392 &  5.426 &   7.463 &  10.180\\ 
8        & 3.624 & 5.005  &   5.459 &  4.817 &   9.948 &  12.963\\ 
9        & 3.634 & 5.510  &   4.882 &  4.798 &   11.160 &  13.605\\ 
10       & 5.283 & 3.731  &   5.142 &  5.450 &   10.753
&  11.884\\ 
\hline \hline
\end{tabular*}
T (s) -- average time in seconds over 10 runs\\Non-Em. -- non-emergency scenario; Em. -- emergency scenario\\
\end{table}

To enhance the efficiency of our framework and prepare it for real-world applications, we leveraged the description of dynamic domains, which specifies elementary actions that cannot physically be executed in certain states (e.g., the agent cannot stop at location $l$ if it is not physically present there).
For this, we extended the signature of \reilp with two predicates: $action\_in\_rule(r, e)$, indicating that the elementary action referenced in rule $r$ is $e$; and $action\_is\_executable(r, i)$, indicating that the action in rule $r$ is physically executable at time step $i$.
 These predicates were defined by the following new rules added to the component \r of \reilp:

$
\begin{array}{lll}
action\_in\_rule(R, E) & \leftarrow & head(R, permitted(E)).\\
action\_in\_rule(R, E) & \leftarrow & head(R, neg(permitted(E))).\\
action\_in\_rule(R, E) & \leftarrow & head(R, obl(E)), action(E).\\
action\_in\_rule(R, E) & \leftarrow & head(R, obl(neg(E))), action(E).\\
action\_in\_rule(R, E) & \leftarrow & head(R, neg(obl(E))), action(E).\\
action\_in\_rule(R, E) & \leftarrow & head(R, neg(obl(neg(E)))), action(E).\\
action\_is\_executable(R, I) & \leftarrow & action\_in\_rule(R, E), \mbox{not } \neg occurs(E, I), step(I).
\end{array}
$

\noindent
We also modified the second and third rules in \r, shown in (\ref{eq:independent_rules}), to require that an action be executable in order for a rule to be deemed applicable (i.e., to hold) at a given time step. The updated rules are encoded as follows:

$
\begin{array}{lll}
    holds(R, I) & \leftarrow & type(R, strict), holds(b(R), I), action\_is\_executable(R, I),\\
    holds(R, I) & \leftarrow & type(R, defeasible), holds(b(R), I), action\_is\_executable(R, I),\\
             & & opp(R, O), \mbox{not } holds(O, I), \mbox{not } holds(ab(R), I)
\end{array}
$

We report the performance results for this improved version of component \r in Table \ref{tbl:asp-core-2}, comparing them against our original framework shown in Table \ref{tbl:performance_traffic}. As shown, performance improves, suggesting that with further enhancements the framework will be closer to practical real-world implementation.

\begin{table}[hbt!]
\caption{Performance Results: Traffic Norms Domain -- Revisited}
\label{tbl:asp-core-2}
\centering
\begin{tabular*}{0.8\linewidth}
{@{\extracolsep{\fill}} c c c c c}
\hline\hline
& \multicolumn{2}{c}{\textbf{Our Improved Framework}}  & \multicolumn{2}{c}{\textbf{Our Original Framework}} \\ \hline
\multicolumn{1}{r}{} & \multicolumn{1}{r}{\textbf{Non-Emergency}} & \multicolumn{1}{r}{\textbf{Emergency}} & \multicolumn{1}{r}{\textbf{Non-Emergency}} & \multicolumn{1}{r}{\textbf{Emergency}}\\
{Scenario \#}    & T (s) & T (s) & T (s) & T (s)\\
\hline
1 & 2.842 & 2.930 & \ \ 3.853 & \ \ 4.108 \\
2 & 2.655 & 2.777 & \ \ 4.494 & \ \ 3.678 \\
3 & 3.447 & 2.695 & \ \ 3.902 & \ \ 3.944 \\
4 & 2.777 & 2.760 & \ \ 4.252 & \ \ 4.682 \\
5 & 2.917 & 2.897 & \ \ 3.644 & \ \ 3.599 \\
6 & 2.725 & 3.077 & \ \ 4.660 & \ \ 3.661 \\
7 & 2.728 & 2.989 & \ \ 4.852 & \ \ 4.837 \\
8 & 2.711 & 3.172  & \ \ 3.624 & \ \ 5.005 \\
9 & 2.649 & 3.042  & \ \ 3.634 & \ \ 5.510 \\
10 & 2.786 & 3.106 & \ \ 5.283 & \ \ 3.731 \\
11 & 5.494 & 5.955 & 18.627 & 12.558 \\
12 & 4.839 & 4.733 & 17.492 & 15.853 \\
13 & 4.622 & 4.528 & 17.085 & 18.223 \\ 
\hline\hline
\end{tabular*}

T (s) -- average time in seconds over 10 runs\\
\end{table}

\section{Related Work}
\label{sec:related_work}

Penalties in the context of norm systems have been discussed by Balke et al. \citeyear{BalkeVP13b,BalkeVP13a}. Shams et al. \citeyear{ShamsVPV17} introduced an ASP framework for reasoning and planning with norms for autonomous agents. As in our work, their actions have an associated duration and can incur penalties. Moreover, their policies have an expiration deadline, which we do not consider in our work. In contrast, our framework leverages policy specification language \aopl, which has built-in means for specifying situations when one policy overrides another one, expressed via defeasible policy rules and \textit{prefer} predicates. Mechanisms for checking \aopl policies for unwanted features such as ambiguity 
have been developed and support the process of producing valid plans. Furthermore, our framework was tested on more complex scenarios than the ones presented by Shams et al., with policies consisting of up to 11 rules as seen in Figure \ref{fig:policies}.  

Other existing work on norms, planning, and penalties tends to assume that pre-generated plans are available to the agent that then needs to choose an optimal plan based on a utility function (penalty or reward) (e.g., \cite{Criado10,Panagiotidi2012ReasoningON,MENEGUZZI2015127}). The BDI architecture \cite{rg91} is the underlying basis of a number of these frameworks (e.g., \cite{Kollingbaum05}). Our framework stems from the AAA \cite{bg08} and AIA \cite{bgb14} architectures. It builds on initial explorations of policy-aware agents \cite{mi21} and prior work on specifying agent behavior modes in relation to policies \cite{hi23,InclezanHT24}.

While a majority of existing approaches focus on agents that are always compliant to norms (e.g., \cite{Oren11,Alechina12}), our intention is to study agent behavior in situations when non-compliant actions must be executed to achieve an important goal (e.g., emergency situations). We believe this to be important to policy makers by simulating human behavior, as humans may act in ways that are non-compliant to cultural norms and even regulations.

\section{Conclusions and Future Work}
\label{sec:conclusions}
In this paper, we introduced a framework for modeling penalties for non-compliant behavior in autonomous agents operating under norm-governed environments. 
Building on the work of Harders and Inclezan \citeyear{hi23}, we refined the distinction between plans containing non-compliant actions, enabling the selection of optimal plans in situations where such behavior is necessary. To this end, we extended \aopl with the ability to associate penalties with policies, capturing different levels of offense severity, and named the extended version \aoplp. 
Additionally, we developed an automated translator from \aoplp into ASP to streamline the use of our framework.
We evaluated our approach in two dynamic domains and found that it generates higher-quality plans while, most importantly, preventing harm to humans, an aspect not addressed in the previous framework.

We identify three key avenues for future work. First, we aim to collaborate with ethics experts to refine the penalty scheme, ensuring it better aligns with societal values and ethical considerations. While our current framework follows a utilitarian approach, alternative ethical perspectives, such as the ethics of care, deserve exploration. Second, we plan to identify and define behavior modes most relevant to policymakers, enabling more realistic simulations of human attitudes toward policies and compliance for policy refinement. Finally, we seek to enhance the efficiency of our framework to improve scalability and performance.

\bibliographystyle{acmtrans.bst}
\bibliography{lpnmr2024}
\end{document}